\documentclass[sigconf,nonacm]{acmart}
\usepackage{algorithm}
\usepackage{algpseudocode}
\usepackage{multirow}
\usepackage[table]{xcolor}
\usepackage{enumitem}
\usepackage{titletoc}
\usepackage{booktabs}
\usepackage{hyperref}
\usepackage[utf8]{inputenc}
\usepackage{tabularx}
\usepackage{array}
\usepackage[dvipsnames, table]{xcolor}
\usepackage{caption}
\usepackage{cleveref}
\usepackage{wrapfig}
\usepackage{float}
\usepackage{pgfplots}
\usepackage{tcolorbox}
\usepackage{flushend}
\pgfplotsset{compat=1.18}
\usepackage{stfloats}

\definecolor{bridge}{HTML}{D35400} 
\definecolor{support}{HTML}{2471A3} 
\definecolor{answer}{HTML}{1E8449} 
\definecolor{general}{HTML}{C000C0} 

\makeatletter
\def\@typeset@author@bx{\bgroup\hsize=\textwidth
  \def\and{\par}\normalbaselines
  \global\setbox\author@bx=\vtop{\centering
    \@authorfont\@currentauthors\par\@affiliationfont
    \@currentaffiliation}\egroup
  \box\author@bx\hspace{\author@bx@sep}%
  \gdef\@currentauthors{}%
  \gdef\@currentaffiliation{}}
\makeatother

\begin{document}

\title{Condition-Gated Reasoning for Context-Dependent Biomedical Question Answering}

\author{%
  \textbf{Jash Parekh}\textsuperscript{$\dagger$}\quad
  \textbf{Wonbin Kweon}\textsuperscript{$\dagger$}\quad
  \textbf{Joey Chan}\textsuperscript{$\dagger$}\quad
  \textbf{Rezarta Islamaj}\textsuperscript{$\diamond$}\quad
  \textbf{Robert Leaman}\textsuperscript{$\diamond$} \\[3pt]
  \textbf{Pengcheng Jiang}\textsuperscript{$\dagger$}\quad
  \textbf{Chih-Hsuan Wei}\textsuperscript{$\diamond$}\quad
  \textbf{Zhizheng Wang}\textsuperscript{$\diamond$}\quad
  \textbf{Zhiyong Lu}\textsuperscript{$\diamond$}\quad
  \textbf{Jiawei Han}\textsuperscript{$\dagger$} \\[7pt]
  $\dagger$~University of Illinois Urbana-Champaign\\
  $\diamond$~National Institutes of Health \\[4pt]
}

\renewcommand{\shortauthors}{Parekh et al.}

\begin{abstract}
Current biomedical question answering (QA) systems often assume that medical knowledge applies uniformly, yet real-world clinical reasoning is inherently conditional: nearly every decision depends on patient-specific factors such as comorbidities and contraindications.
Existing benchmarks do not evaluate such conditional reasoning, and retrieval-augmented or graph-based methods lack explicit mechanisms to ensure that retrieved knowledge is applicable to given context.
To address this gap, we propose \textbf{CondMedQA}, the first benchmark for conditional biomedical QA, consisting of multi-hop questions whose answers vary with patient conditions.
Furthermore, we propose \textbf{\textsc{Condition-Gated Reasoning}} \textsc{(CGR)}, a novel framework that constructs condition-aware knowledge graphs and selectively activates or prunes reasoning paths based on query conditions.
Our findings show that CGR more reliably selects condition-appropriate answers while matching or exceeding state-of-the-art performance on biomedical QA benchmarks, highlighting the importance of explicitly modeling conditionality for robust medical reasoning.
\end{abstract}




\maketitle

\begin{figure}[t]
    \centering
    \includegraphics[width=1\linewidth]{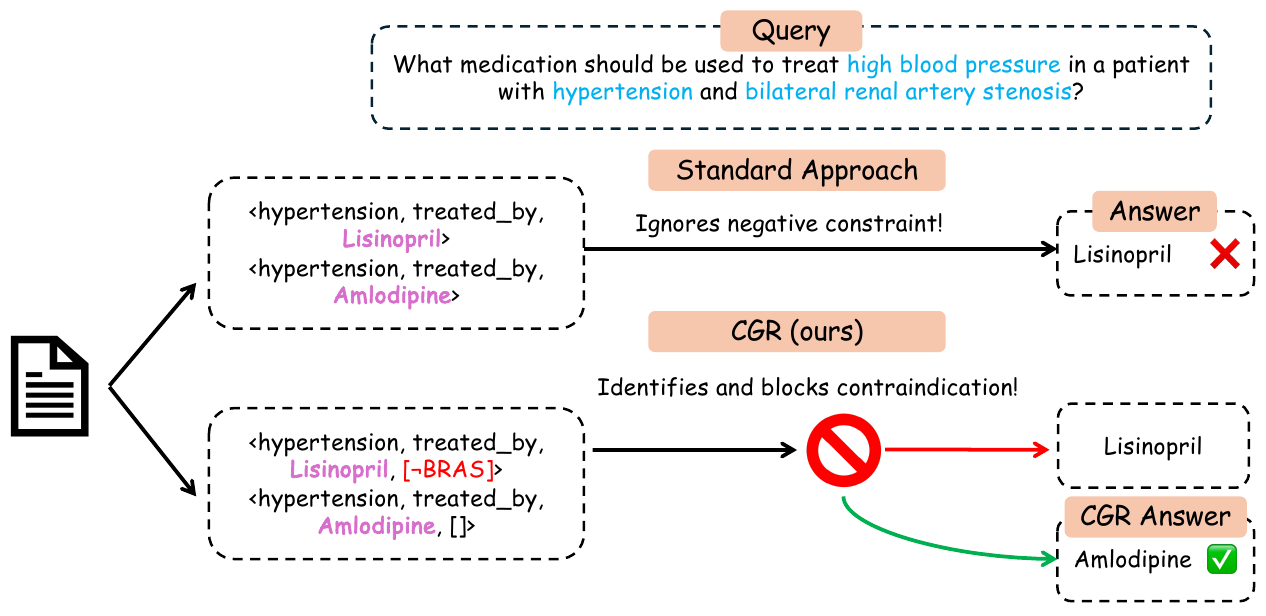}
    \caption{\normalfont Example of condition-gated reasoning in the biomedical domain. Existing KG-RAG extracts triples and retrieves contraindicated treatments. CGR extracts n-tuples with patient-specific conditions, gating {\color{red}unsafe paths} and retrieving only {\color{green!50!black}safe alternatives}.}
    \label{fig:intro_fig}
\end{figure}

\section{Introduction}

Retrieval-augmented generation (RAG) \cite{lewis2020retrieval} has emerged as an effective paradigm for grounding the reasoning process of large language models (LLMs) in external knowledge bases, reducing hallucinations and enabling access to up-to-date information.
To overcome the limitations of standard RAG in multi-hop reasoning, recent approaches (e.g., GraphRAG \cite{edge2024}, HippoRAG \cite{gutierrez2024}) have introduced graph-structured retrieval.
In this setting, knowledge is organized as graphs whose fundamental unit is a triplet of $\langle \text{entity}, \text{relation}, \text{entity} \rangle$ capturing explicit semantic relationships.
In the biomedical domain, recent approaches (e.g., MedRAG \cite{medrag2024} and MKRAG \cite{shi2025mkrag}) build biomedical domain-specific knowledge graphs from medical documents and electronic health records (EHRs).
This structured biomedical knowledge enables more reliable and interpretable reasoning, allowing LLMs to generate clinically-grounded answers.

While knowledge graph-based approaches effectively capture relational structure, they typically encode only binary relationships and do not explicitly model clinical preferences or how these preferences shift under specific conditions. Biomedical question answering is inherently conditional, as clinical decision-making requires reasoning over patient-specific contexts such as allergies, comorbidities, and concurrent medications to choose the preferred action (or actions) given the overall clinical picture. For instance, lisinopril is commonly used as a first-line treatment for hypertension. However, lisinopril, as an ACE inhibitor, is contraindicated in patients with bilateral renal artery stenosis, and amlodipine, a calcium channel blocker, becomes a preferred alternative. Conventional knowledge graphs may encode these facts as $\langle \text{lisinopril}, \allowbreak \text{treats}, \allowbreak \text{hypertension} \rangle$ and $\langle \mathrm{amlodipine}, \allowbreak \mathrm{treats}, \allowbreak \mathrm{hypertension} \rangle$, but such representations do not capture the default preference for lisinopril, nor how this preference shifts under specific constraints, even if the knowledge graph explicitly records $\mathopen{\langle}
\text{lisinopril},\allowbreak
\text{contraindicated in},\allowbreak
\text{bilateral renal artery stenosis}
\mathclose{\rangle}$. Retrieved passages often describe treatments, contraindications, or preferences in isolation, without making explicit how these considerations interact in a specific patient context. As a result, a system must determine not only which facts are relevant, but which combinations of facts jointly apply, and whether missing contextual information invalidates a candidate answer.

\smallskip

\noindent\textbf{Contribution 1: Benchmark for conditional biomedical QA.}
The absence of condition-aware reasoning is also reflected in existing evaluation sets.
Established biomedical QA benchmarks, such as MedHopQA \cite{islamaj2026overview}, MedHop \cite{welbl2018constructing}, BioCDQA \cite{feng2025retrieval}, and BioASQ \cite{tsatsaronis2015overview}, primarily focus on factual recall and multi-hop reasoning.
These benchmarks do not systematically evaluate whether approaches can modulate their responses based on contextual constraints. 
Our work addresses this identified gap, by introducing \textbf{CondMedQA}, a diagnostic benchmark comprising of 100 curated questions designed to evaluate conditional reasoning in biomedical QA. To the best of our knowledge, CondMedQA represents the first benchmark specifically targeting these capabilities. Each question requires identification of the correct response given explicit constraints that modify the applicability of standard knowledge. 

\smallskip

\noindent\textbf{Contribution 2: Framework for conditional biomedical QA.}
Furthermore, we propose \textbf{\textsc{Condition-Gated Reasoning}} \textsc{(CGR)}, a framework that prioritizes conditional representation within knowledge graph construction and traversal. At a high level, CGR treats conditions not as attributes to be inferred implicitly by a language model, but as explicit validity constraints on retrieved knowledge.
Rather than aggregating evidence solely based on semantic relevance, CGR enforces compatibility between patient-specific context and the conditions under which medical relationships hold, ensuring that only applicable information contributes to downstream reasoning.
This design enables structured multi-hop inference in which intermediate conclusions are filtered by conditional constraints, preventing the propagation of contraindicated 
or inapplicable facts.
As a result, CGR is specifically suited to biomedical queries in which correct answers emerge only through the simultaneous satisfaction of multiple interacting conditions.

\smallskip

We evaluate CGR on CondMedQA and several established biomedical QA benchmarks, comparing against state-of-the-art reasoning methods.
Our key contributions are summarized as follows.

\begin{itemize}[leftmargin=10pt, itemsep=2pt]
    \item CondMedQA: a new benchmark that enables systematic evaluation of conditional multi-hop reasoning in the biomedical domain, requiring models to account for patient-specific constraints when determining the applicability of medical knowledge.
    \item \textsc{Condition-Gated Reasoning}: a novel framework that explicitly models condition-aware constraints within knowledge graph construction and traversal, ensuring that only contextually valid information contributes to multi-hop inference.
\end{itemize}

We demonstrate the capabilities of CGR through extensive experiments on CondMedQA and other benchmarks, achieving substantial gains on condition-sensitive queries while matching or exceeding state-of-the-art performance on factual benchmarks.

\section{Related Work}

\smallskip

\noindent \textbf{Reasoning Limitations in Standard RAG.}
Although Retrieval-Augmented Generation (RAG) provides LLMs with external evidence, conventional frameworks are often hampered by their reliance on the model's ability to internally bridge gaps across unstructured context. Research has highlighted the ``lost-in-the-middle'' effect \cite{liu2024, jiang2025retrieval, jiang2025ras} and fragmentation of information \cite{jiang2024longrag}, which prevent models from recognizing logical links across documents. While strategies like long-context fine-tuning \cite{xiong2024effective} and memory-augmented retrieval \cite{qian2025memorag} attempt to mitigate these issues, they do not address the difficulty of reasoning over unstructured data.

\smallskip

\noindent \textbf{Knowledge Graph-based Reasoning.}
One emerging approach involves the pre-emptive construction of comprehensive knowledge graphs across an entire document collection prior to the inference stage. For instance, GraphRAG \cite{edge2024} organizes a full corpus into hierarchical community structures and utilizes pre-computed summaries to facilitate global-scale QA. Similarly, LightRAG \cite{guo2024} maps relationships between entities and text segments, requiring a thorough traversal of the entire dataset during indexing. HippoRAG \cite{gutierrez2024} maintains a global graph state and employs Personalized PageRank to enable associative retrieval. HyperGraphRAG \cite{luo2025} introduces hypergraph architectures to represent higher-order dependencies between entities and documents, supporting more complex multi-hop reasoning over large-scale data. PathRAG \cite{chen2025pathrag} prunes retrieved subgraphs by identifying reasoning paths most relevant to the query, reducing noise from loosely connected edges. SARG \cite{parekh2025sarg} synthesizes knowledge graphs from retrieved passages with bidirectional traversal to explicitly surface multi-hop reasoning chains.

\begin{table*}[t]
\centering
\footnotesize
\caption{\normalfont Types of conditional multi-hop reasoning required in CondMedQA. We show \textcolor{bridge}{\textbf{\textit{orange bold italics}}} for bridge entities, \textcolor{support}{\textit{blue italics}} for supporting facts, \underline{underline} for the conditions, \textcolor{answer}{\textbf{green bold}} for the conditional answer, and \textcolor{general}{magenta} for the general (non-conditional) answer. Each example requires synthesizing information across both documents, neither document alone contains sufficient information to answer the conditional question.}
\label{tab:examples}

\renewcommand{\arraystretch}{1.1}
\setlength{\tabcolsep}{6pt}

\begin{tabularx}{\textwidth}{>{\raggedright\arraybackslash}p{4.5cm} c X}
\toprule
\textbf{Reasoning Category} & \textbf{\%} & \textbf{Example(s)} \\
\midrule
Comorbidity \& Organ-Based Contraindications & 57 & 
\textbf{Doc 1:} Treatment for \textcolor{support}{\textit{major depressive disorder}} includes \textcolor{support}{\textit{antidepressants}} such as \textcolor{answer}{\textbf{bupropion}}, \textcolor{general}{sertraline}, and \textcolor{support}{\textit{fluoxetine}}. \textcolor{bridge}{\textbf{\textit{Bupropion}}} acts as a norepinephrine-dopamine reuptake inhibitor. \\
& & \textbf{Doc 2:} \textcolor{bridge}{\textbf{\textit{Bupropion}}} lowers the seizure threshold and is contraindicated in patients with \textcolor{support}{\textit{bulimia nervosa}} or other \textcolor{support}{\textit{eating disorders}} due to elevated seizure risk. \\
& & \textbf{Q:} Which antidepressant for MDD is contraindicated in patients \underline{with bulimia}? \\
& & \textbf{A:} \textcolor{answer}{\textbf{Bupropion}} (w/o condition: \textcolor{general}{Sertraline}) \\
\midrule

Diagnostic Modality Selection & 23 & 
\textbf{Doc 1:} For suspected \textcolor{support}{\textit{appendicitis}}, diagnostic imaging options include \textcolor{general}{CT scan}, \textcolor{answer}{\textbf{ultrasound}}, and \textcolor{support}{\textit{MRI}}. \textcolor{support}{\textit{CT scan}} has the highest sensitivity in adult populations. \\
& & \textbf{Doc 2:} In \textcolor{support}{\textit{pediatric patients}}, \textcolor{bridge}{\textbf{\textit{ultrasound}}} is the preferred first-line imaging modality to avoid \textcolor{support}{\textit{ionizing radiation exposure}}, with CT reserved for inconclusive cases. \\
& & \textbf{Q:} What imaging test is preferred for suspected appendicitis \underline{in children}? \\
& & \textbf{A:} \textcolor{answer}{\textbf{Ultrasound}} (w/o condition: \textcolor{general}{CT Scan}) \\
\midrule

Special Population Safety & 16 & 
\textbf{Doc 1:} First-line antibiotics for \textcolor{support}{\textit{Lyme disease}} include \textcolor{general}{doxycycline}, \textcolor{answer}{\textbf{amoxicillin}}, and \textcolor{support}{\textit{cefuroxime}}. \textcolor{bridge}{\textbf{\textit{Doxycycline}}} is preferred in adults due to co-coverage of \textcolor{support}{\textit{Anaplasma}} co-infection. \\
& & \textbf{Doc 2:} \textcolor{bridge}{\textbf{\textit{Doxycycline}}} is contraindicated during \textcolor{support}{\textit{pregnancy}} due to risks to fetal \textcolor{support}{\textit{bone and teeth development}}. \textcolor{answer}{\textbf{Amoxicillin}} is considered safe across all trimesters. \\
& & \textbf{Q:} What antibiotic is recommended for Lyme disease in a \underline{pregnant patient}? \\
& & \textbf{A:} \textcolor{answer}{\textbf{Amoxicillin}} (w/o condition: \textcolor{general}{Doxycycline}) \\
\midrule

Drug-Drug Interactions \& Pharmacogenomics & 4 & 
\textbf{Doc 1:} Standard therapy for \textcolor{support}{\textit{tuberculosis}} includes \textcolor{general}{rifampin}, \textcolor{support}{\textit{isoniazid}}, \textcolor{support}{\textit{pyrazinamide}}, and \textcolor{support}{\textit{ethambutol}}. \textcolor{bridge}{\textbf{\textit{Rifabutin}}} is an alternative rifamycin with a similar mechanism. \\
& & \textbf{Doc 2:} \textcolor{general}{Rifampin} is a potent inducer of CYP3A4 and is contraindicated with \textcolor{support}{\textit{HIV protease inhibitors}}. \textcolor{bridge}{\textbf{\textit{Rifabutin}}} has fewer CYP3A4 interactions and is preferred in patients on \textcolor{support}{\textit{antiretroviral therapy}}. \\
& & \textbf{Q:} Which drug replaces rifampin in TB treatment for \underline{HIV patients on protease inhibitors}? \\
& & \textbf{A:} \textcolor{answer}{\textbf{Rifabutin}} (w/o condition: \textcolor{general}{Rifampin}) \\
\bottomrule
\end{tabularx}
\end{table*}

\smallskip

\noindent \textbf{Reasoning for Biomedical QA.}
Advances in biomedical reasoning have led to specialized RAG frameworks such as MedRAG \cite{medrag2024}, MedGraphRAG \cite{wu2024medical}, KRAGEN \cite{matsumoto2024kragen}, and MKRAG \cite{shi2025mkrag} which move beyond flat text retrieval by integrating domain-specific structure.
Specifically, approaches like MedRAG construct specialized knowledge graphs from medical literature and electronic health records (EHRs), connecting entities such as drugs, diseases, and genes through clinically relevant relations.
By leveraging these structured representations, these methods elicit more reliable and interpretable reasoning, allowing LLMs to ground diagnosis and treatment recommendations in observed clinical manifestations.
To evaluate these capabilities, established benchmarks such as MedHop \cite{welbl2018constructing}, BioCDQA \cite{feng2025retrieval}, BioASQ \cite{tsatsaronis2015overview}, PubMedQA \cite{jin2019pubmedqa}, and BioHopR \cite{biohopr2024} have become standard for measuring factual recall and multi-hop reasoning in the healthcare domain.

\smallskip

\noindent \textbf{Non-Monotonic Reasoning.}
Monotonic logic assumes that once a conclusion is derived, it remains valid as new premises are added. In contrast, clinical reasoning is inherently non-monotonic: a previously valid recommendation may be overridden by new patient information, such as contraindications or drug interactions. Formal frameworks for non-monotonic reasoning include Reiter's default logic~\cite{reiter1980logic}, McCarthy's circumscription~\cite{mccarthy1980circumscription}, stable model semantics~\cite{fitting1992michael}, and Dung's argumentation frameworks~\cite{dung1995acceptability}. These frameworks have been explored in medical decision support~\cite{fox2000safe}, but rule-based systems often struggle with incomplete knowledge and the combinatorial complexity that come from multiple co-occurring conditions and medications.

\section{CondMedQA Benchmark}
\label{sec:benchmark}


Existing biomedical QA benchmarks evaluate factual recall or multi-hop reasoning, but do not systematically evaluate \textit{conditional reasoning}, i.e. the ability to modulate answers based on patient-specific constraints (illustrated in Fig.~\ref{fig:pipeline}).
To address this gap, we introduce CondMedQA, a diagnostic benchmark of 100 questions designed to probe this capability in current state-of-the-art approaches.
In this section, we first formalize the conditionality in biomedical QA, and elaborate our construction process of CondMedQA.
Table~\ref{tab:examples} presents multiple types of examples from CondMedQA benchmark. 

\subsection{Formalization of Conditionality} 
\label{subsec:formalization}

We define a question as \emph{conditional} if and only if it satisfies the following criteria:
\begin{enumerate}[leftmargin=*, itemsep=2pt, topsep=2pt]
    \item \textbf{Modifier presence}: The question contains a specific patient condition (e.g., pregnancy, comorbidity, genetic factor).
    \item \textbf{General answer existence}: There exists a standard or default answer that would apply if the modifier were absent.
    \item \textbf{Answer divergence}: The correct answer given the modifier differs from the general answer.
    \item \textbf{Dual validity}: The general answer is correct in typical cases, conditional answer is correct specifically due to the modifier.
    \item \textbf{Causal dependence}: The modifier directly causes the answer to change, removing the modifier reverts the to the general case.
\end{enumerate}
\noindent This formalization ensures that LLMs cannot succeed by memorizing default answers and must recognize how patient-specific constraints adjust the applicability of medical knowledge.

\subsection{Data Construction}
\label{subsec:construction}

We construct CondMedQA through a three-stage pipeline designed to balance question diversity with quality.

\subsubsection{Candidate Generation}
We prompt Gemini-3-Pro \cite{team2023gemini} with few-shot examples of conditional medical questions, instructing it to generate question-answer pairs where patient-specific factors change the correct response using two Wikipedia articles. The model is prompted to provide: (1) the conditional question, (2) the correct answer given the condition, (3) what the general answer would be without the condition, (4) an explanation of why the condition changes the answer, and (5) the two Wikipedia articles used to synthesize the question and answer. This structured output facilitates downstream verification.

\subsubsection{Automated Filtering} Generated candidates undergo rule-based filtering to remove: (1) questions where the ``conditional'' and ``general'' answers are identical, (2) questions lacking explicit patient modifiers, (3) duplicates via embedding similarity, (4) questions with ambiguous or multiple valid answers. and (5) questions with invalid Wikipedia links.

\subsubsection{Manual Verification} All candidates underwent review by members of our research team. For each question, reviewers verify: (1) the question satisfies all five conditionality criteria, (2) the provided answer is consistent with medical references, (3) the question is unambiguous and well-formed, and (4) both Wikipedia documents are used to form a logical reasoning trace that contains the answer. Reviewers correct phrasing issues and reject questions that fail verification.
\subsection{Quality Assurance}
\label{subsec:quality}

To further validate quality, a subset of 30 questions underwent independent review by three annotators with medical expertise. Annotators evaluate each question on three dimensions: (1) \textit{Conditionality} (Yes/No), whether the question satisfies all criteria in \S\ref{subsec:formalization}; (2) \textit{Answer Accuracy} (Correct/Incorrect/Uncertain), whether the provided answer aligns with clinical guidelines; and (3) \textit{Question Quality} (1--5), overall clarity, clinical relevance, and answerability. Inter-annotator agreement is reported in Table~\ref{tab:iaa}. Gwet's AC1 (nominal) and AC2 (ordinal) \cite{gwet2001handbook} are reported as the primary chance-corrected metrics, as Krippendorff's $\alpha$ \cite{krippendorff2011computing} and Cohen's $\kappa$ \cite{cohen1960coefficient} are unreliable under highly skewed marginal distributions ~\citep{gwet2008computing}. Note that \textit{\% Agree (all)} requires exact agreement across all three annotators (e.g., all rate quality as 5), while \textit{\% Agree (pair)} reflects pairwise agreement; the latter better captures consensus given the inherent subjectivity of quality judgments.

\begin{table}[H]
\centering
\caption{\normalfont Inter-annotator agreement across 30 items.}
\label{tab:iaa}
\small
\begin{tabular}{lccc}
\toprule
\textbf{Dimension} & \textbf{Gwet's AC} & \textbf{\% Agree (all)} & \textbf{\% Agree (pair)} \\
\midrule
Conditionality & 0.96 (AC1) & 96.7 & 97.8 \\
Answer Accuracy & 0.67 (AC1) & 66.7 & 77.8 \\
Question Quality & 0.60 (AC2) & 56.7 & 71.1 \\
\bottomrule
\end{tabular}
\end{table}


\subsection{Question Types}
\label{subsec:question_types}

We categorize the conditional reasoning patterns in CondMedQA into four types based on what modifies the clinical decision:

\begin{enumerate}
    \item \textbf{Comorbidity \& Organ-Based Contraindications (57\%).} Encompasses disease contraindications (e.g., avoiding beta-blockers in asthma) and organ dysfunction requiring dosage adjustment or drug substitution.
    \item \textbf{Diagnostic Modality Selection (23\%).} Questions where the answer is an imaging test rather than a drug, requiring reasoning about radiation exposure, contrast contraindications, or device compatibility (e.g. MRI-unsafe pacemakers). 
    \item \textbf{Special Population Safety (16\%).} Life-stage considerations including pregnancy safety categories, pediatric dosing constraints, and geriatric precautions.
    \item \textbf{Drug-Drug Interactions \& Pharmacogenomics (4\%).} Most specific category involving enzyme interactions, contraindicated drug combinations, and genetic variations.
\end{enumerate}

\noindent Table~\ref{tab:examples} provides detailed examples of each reasoning type, illustrating how information from two documents must be synthesized to arrive at the correct conditional answer.


\begin{figure*}[t]
    \centering
    \includegraphics[width=1\linewidth]{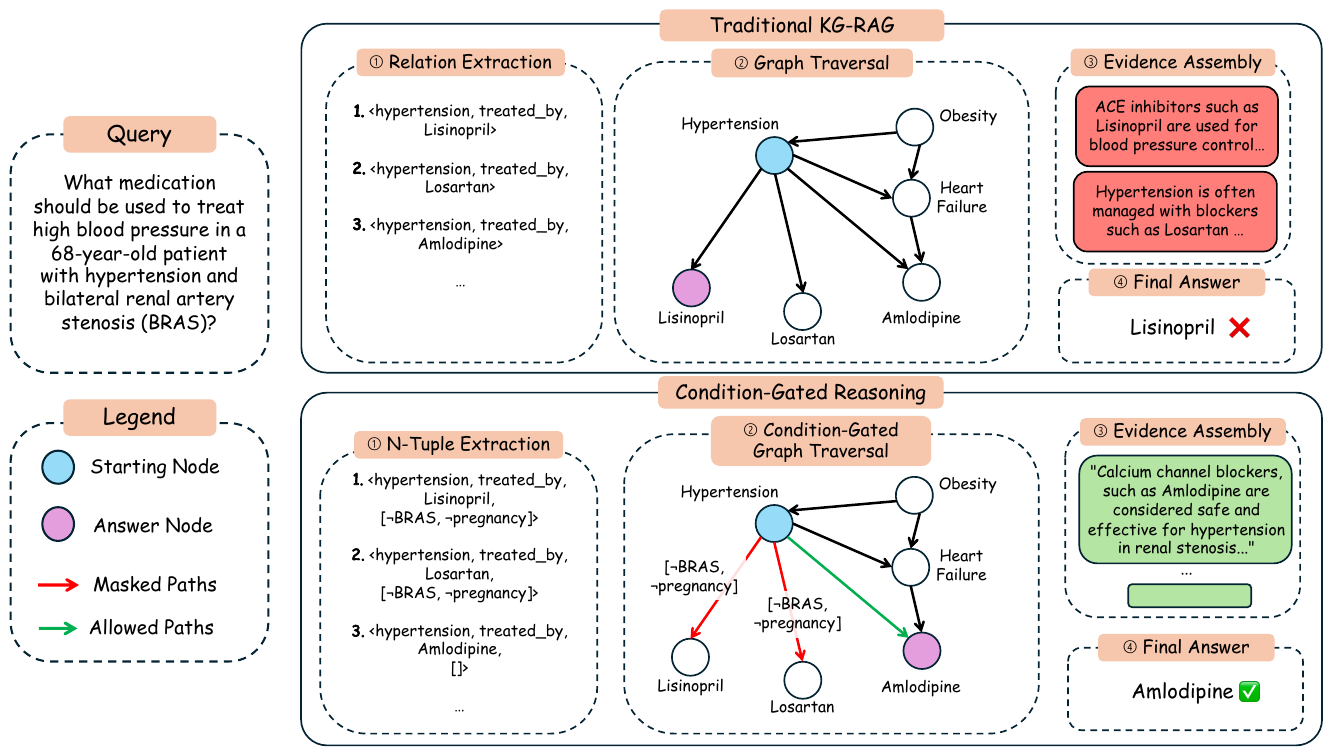}
    \caption{\normalfont Overview of \textsc{Condition-Gated Reasoning (CGR)} compared to traditional KG-RAG. Given a clinical query about a patient with hypertension and bilateral renal artery stenosis (BRAS), traditional KG-RAG extracts standard relation triples and traverses all paths indiscriminately, retrieving evidence for contraindicated treatments (e.g., Lisinopril, an ACE inhibitor). CGR extends triples to n-tuples that include patient-specific conditions as gating constraints (e.g., $\neg$BRAS, $\neg$pregnancy), masking contraindicated paths during graph traversal and assembling only condition-appropriate evidence, yielding the correct answer (Amlodipine).}
    \label{fig:pipeline}
\end{figure*}

\section{Condition-Gated Reasoning}
\label{sec:cgr}
We propose \textbf{\textsc{Condition-Gated Reasoning}} \textsc{(CGR)}, a framework that prioritizes conditional representation within knowledge graph construction and traversal.
CGR addresses conditional biomedical QA by explicitly representing contextual conditions as integral components of knowledge graph edges.
Unlike standard graph-based retrieval methods that traverse all edges uniformly, CGR gates edge traversal based on compatibility between edge conditions and query context.
Figure~\ref{fig:pipeline} illustrates the complete pipeline.


\subsection{Condition-Aware Knowledge Graph}
\label{subsec:graph}
In this subsection, we describe the construction of a condition-gated biomedical knowledge graph that encodes not only entity–entity relations but also the clinical conditions under which they apply.

\subsubsection{Condition-Aware Tuple Extraction}
\label{subsec:extraction}

Given a corpus of source documents $\mathcal{D}$, we extract structured 4-tuples of the form $\langle u, r, v, \mathcal{C} \rangle$, 
where $u$ and $v$ are nodes, $r$ is a relation, and $\mathcal{C} = [c_1, \ldots, c_k]$ is a list of conditions under which the relationship holds or does not hold.
Conditions $\mathcal{C}$ capture contextual constraints including patient demographics (e.g., ``pediatric patients''), physiological status (e.g., ``during pregnancy''), comorbidities (e.g., ``with renal impairment''), disease stage (e.g., ``early localized''), and contraindication contexts (e.g., ``penicillin allergy''). Extraction is performed using Qwen2.5-14B-Instruct-GPTQ-Int4~\cite{team2024qwen2}. Examples of the extracted tuples and associated conditions are provided in Appendix~\ref{app:extraction_examples}.

\subsubsection{Entity Normalization}
\label{subsec:normalization}
Extracted entities are normalized to canonical forms using the UMLS Metathesaurus~\cite{schuyler1993umls}, ensuring consistent representation across synonymous terms (e.g., ``heart attack'' $\rightarrow$ ``myocardial infarction'').

\subsubsection{Gated Knowledge Graph Construction}
Normalized tuples are assembled into a directed knowledge graph $G = (V, E)$, where nodes $v \in V$ correspond to unique entities and edges $e \in E$ encode relationships with associated conditions.
Each edge $e = \langle u, r, v, \mathcal{C} \rangle$ carries $S_e$, a set of evidence snippets that provide supporting text spans.
Our knowledge graph structure permits multiple edges between the same node pair with distinct relations or conditions.
\subsection{Query Processing}
\label{subsec:query}
This subsection describes how queries are converted into structured forms for graph-based reasoning.
We parse queries to extract entities and conditions, and use LLM-based evaluation to align query semantics with condition-aware graph edges.

\subsubsection{Query Parsing}
Given a natural language query $q$, we extract a structured representation: $\text{parse}(q) = \{K, C, N\}$
where $K$ is a set of entity keywords, $C$ represents required and excluded conditions, and $N$ are negated entities to exclude from candidate answers. Query parsing is performed using Qwen2.5-14B-Instruct-GPTQ-Int4 \cite{team2024qwen2}. Examples are provided in Table~\ref{tab:query_parsing_examples} (Appendix~\ref{app:query_parsing}).

\subsubsection{LLM-Based Condition Evaluation}
\label{subsec:condition_eval}
A key challenge in condition matching is semantic variability: the query may express conditions differently than the graph edges (e.g., ``5-year-old patient'' vs.\ ``in children''; ``kidney disease'' vs.\ ``renal impairment''). Traditional keyword matching fails on synonyms and negations. 

\noindent We address this through LLM-based condition evaluation. Prior to graph traversal, we collect all unique conditions $\mathcal{C}_G = \bigcup_{e \in E} \mathcal{C}_e$ from the graph and evaluate them against the query in a single LLM call: $\text{eval}: \mathcal{C}_G \times q \rightarrow \{\texttt{True}, \texttt{False}, \texttt{Null}\}^{|\mathcal{C}_G|}$. The evaluation returns \texttt{True} if the query context satisfies the condition, \texttt{False} if it explicitly violates the condition, and \texttt{Null} if the query provides no relevant information. This single-pass evaluation yields a lookup table $\mathcal{L}: \mathcal{C}_G \rightarrow \{\texttt{True}, \texttt{False}, \texttt{Null}\}$, enabling O(1) condition checks during traversal. Examples can be found in Table~\ref{tab:condition_eval_examples} (Appendix~\ref{app:condition_eval}).

\subsection{Reasoning over Condition-Gated Graph}
This subsection presents our framework for traversing and reasoning over the knowledge graph (Algorithm~\ref{alg:gating}).

\subsubsection{Condition-Gated Graph Traversal} \label{subsubsec:traversal}
Given a process query, we describe how CGR traverses the constructed knowledge graph, where edges are gated by conditions.

\smallskip

\noindent \textbf{Entry Node Selection.} 
\label{subsubsec:entry}
We identify entry nodes via semantic matching between query entities and graph nodes.
Specifically, we encode both the query and all node labels using MedEmbed~\cite{balachandran2024medembed}, then select the top-k nodes (ablated in \S\ref{subsec:ablation}) whose cosine similarity exceeds a threshold $\tau$ as starting points.

\smallskip

\noindent \textbf{Edge Gating.}
We perform breadth-first search from entry nodes, where edge traversal is gated by the condition lookup table $\mathcal{L}$. For an edge $e = \langle u, r, v, \mathcal{C}_e \rangle$ with conditions $\mathcal{C}_e = \{c_1, \ldots, c_k\}$, we define a gating function:
\begin{equation}
\mathcal{G}(e, \mathcal{L}) = \prod_{c \in \mathcal{C}_e} \mathbb{1}[\mathcal{L}(c) \neq \texttt{false}]
\end{equation}
where $\mathbb{1}[\cdot]$ is the indicator function. An edge is traversable ($\mathcal{G} = 1$) only if \textit{no} condition evaluates to \texttt{false}. Conditions evaluating to \texttt{null} (unknown) do not block traversal. This conservative policy prunes edges only when the query explicitly violates a condition, and the resulting subgraph $G^{(q)}$ contains only edges satisfying patient context.

\smallskip

\noindent \textbf{Termination Criteria.}
Traversal along a path terminates when: (1) a maximum depth $d_{\max}$ is reached, (2) no outgoing edges satisfy the gating condition, or (3) the node has no outgoing edges (leaf node). This allows multi-hop reasoning chains of up to $d_{\max}$ edges, where each hop is gated by patient context.
All paths that are collected into the set of candidate reasoning paths \( \mathcal{P}_q \).

\smallskip

\noindent \textbf{Example.}
Consider the query in Figure \ref{fig:pipeline}: ``What medication for hypertension in a 68-year-old patient with bilateral renal artery stenosis (BRAS)?''.
Starting from node `hypertension', the traversal encounters edges to various treatments:
\begin{itemize}[leftmargin=15pt, itemsep=1pt, topsep=2pt]
    \item $\langle$hypertension, treated\_by, lisinopril, [$\neg$BRAS, $\neg$pregnancy]$\rangle$ \\ — \textbf{blocked}, since $\mathcal{L}(\text{$\neg$BRAS}) = \texttt{false}$
    \item $\langle$hypertension, treated\_by, losartan, [$\neg$BRAS]$\rangle$ \\ — \textbf{blocked}, since $\mathcal{L}(\neg\text{BRAS}) = \texttt{false}$
    \item $\langle$hypertension, treated\_by, amlodipine, []$\rangle$ \\ — \textbf{traversed}, no conditions violated
\end{itemize}
The gating mechanism prunes contraindicated paths, leaving only safe treatment options for context assembly.

\begin{algorithm}[t]
\begin{algorithmic}[1]
\Require Query $q$, Graph $G=(V,E)$ with edge conditions $C_e$
\Ensure Lookup dict $\mathcal{L}$, Traversable subgraph $G'$
\State $\mathcal{C} \gets \bigcup_{e \in E} C_e$ \Comment{Collect unique conditions}
\State $\mathcal{L} \gets \Call{LLM-Evaluate}{q, \mathcal{C}}$ \Comment{Single LLM call}
\For{each edge $e = (u, v)$ during BFS}
    \For{each $c \in C_e$}
        \If{$\mathcal{L}[c] = \texttt{false}$}
            \State \textbf{skip} $e$ \Comment{Query contradicts condition}
        \EndIf
    \EndFor
    \State Add $e$ to $G'$; enqueue $v$
\EndFor
\end{algorithmic}
\caption{Condition-Gated Edge Filtering}
\label{alg:gating}
\end{algorithm}

\subsubsection{Path Ranking and Answer Generation} \label{subsec:ranking}
CGR then ranks the resulting reasoning paths, and generates grounded answers from the top-ranked evidence.

\smallskip

\noindent \textbf{Path Ranking.}
Candidate paths $p \in \mathcal{P}_q$ are ranked by aggregate semantic similarity between the 
path embedding and query keywords $k \in K$:
\begin{equation}
\label{eq:path-ranking}
\text{score}(q, p) = \sum_{k \in K} \cos\!\Big(\phi(p),\, \phi(k)\Big)
\end{equation}
where $\phi(\cdot)$ denotes the MedEmbed \cite{balachandran2024medembed} embedding function
and $\phi(p)$ is the embedding of the full linearized path
$\langle e_1, r_1, e_2, r_2, \ldots, e_n \rangle$ concatenated as a text sequence.
The top-\(N\) paths \( \mathcal{P}_q^N \) are selected as evidence for answer generation.

\smallskip

\noindent \textbf{Evidence Assembly.} 
Selected paths \( \mathcal{P}_q^N \) are assembled into a structured evidence package $\mathcal{E}_q$ for answering query $q$:
\begin{equation}
\mathcal{E}_q = \{(V_p, E_p, S_p, C_p)\}_{p \in \mathcal{P}_q^N }
\end{equation}
where \( V_p \) denotes the sequence of entities along path \( p \), \( E_p \) the set of edges, \( S_p \) the source text snippets supporting the edges, and \( C_p \) the conditions associated with the traversed edges.
This structured format preserves each path and enables the model to trace reasoning back to source documents.

\smallskip

\noindent \textbf{Answer Generation.} The final answer is generated by passing a prompt comprising of the query, top-$k$ reasoning paths, and their associated evidence to an LLM.
\begin{equation}
A = \text{LLM}\Big(\underbrace{q \oplus \mathcal{P}^N_q \oplus \mathcal{E}_q \oplus \texttt{instructions}}_{\text{prompt}}\Big)
\end{equation}
The instructions direct the model to synthesize a grounded response while respecting the conditions that gated traversal.

\subsubsection{Computational Complexity}
\label{subsec:complexity}

CGR requires $O(1)$ LLM calls for condition evaluation regardless of graph size, as all conditions are evaluated in a single batch. Graph traversal is $O(|V| + |E|)$ with $O(1)$ condition lookup per edge. The primary computational cost is tuple extraction, which scales linearly with corpus size.

\section{Experiments}
\label{sec:experiments}

We evaluate CGR against retrieval-augmented baselines across four biomedical QA benchmarks.

\subsection{Experimental Setup}
\label{subsec:setup}

\subsubsection{Datasets}
We evaluate on four benchmarks spanning factoid and multi-hop biomedical QA:
\begin{enumerate}[leftmargin=*, itemsep=2pt, topsep=2pt]
    \item \textbf{CondMedQA} (ours): 100 questions requiring conditional reasoning over patient-specific constraints (\S\ref{sec:benchmark}).
    \item \textbf{MedHopQA} \cite{islamaj2026overview}: 400 clinically reviewed multi-hop questions requiring reasoning across multiple medical documents.
    \item \textbf{MedHopQA (Cond)}: A disjoint, clinically reviewed set of 35 conditional questions drawn from the same MedHopQA source but excluded from the 400-question split above.
    \item \textbf{BioASQ Task B} \cite{tsatsaronis2015overview}: Biomedical factoid questions. We randomly sample 500 questions from the full test set (5{,}389 questions) due to the computational cost of LLM-based graph construction and evaluation.
\end{enumerate}

\begin{table*}[t]
\centering
\caption{\normalfont Performance comparison across biomedical multi-hop QA datasets and our custom benchmark. \textbf{Bold} indicates the best performance within each foundation model group.}
\label{tab:main_results}
\small
\setlength{\tabcolsep}{16pt}
\begin{tabular}{l|cc|cc|cc|cc}
\toprule
\multirow{2}{*}{\textbf{Method}} 
& \multicolumn{2}{c|}{\textbf{CondMedQA}} 
& \multicolumn{2}{c|}{\textbf{MedHopQA}} 
& \multicolumn{2}{c|}{\textbf{MedHopQA (Cond)}} 
& \multicolumn{2}{c}{\textbf{BioASQ-B}} \\
\cmidrule(lr){2-3} \cmidrule(lr){4-5} \cmidrule(lr){6-7} \cmidrule(lr){8-9}
 & EM & F1 & EM & F1 & EM & F1 & EM & F1 \\
\midrule
 \multicolumn{9}{c}{\cellcolor{gray!10}\textit{GPT-5.2}} \\
\midrule
Zero-Shot & 22.00 & 41.29 & 20.50 & 47.87 & 28.27 & 53.52 & 13.00 & 40.21 \\
RAG & 44.00 & 52.36 & 51.25 & 62.61 & 45.71 & 55.04 & 31.80 & 54.80 \\
CoT & 40.00 & 52.81 & 49.50 & 57.55 & 50.00 & 59.38 & 25.60 & 47.81 \\
HippoRAG2 & 46.00 & 67.56 & 45.00 & 60.81 & 48.57 & 72.82 & 22.60 & 43.57 \\
HyperGraphRAG & 57.00 & 64.48 & 37.50 & 41.90 & 54.29 & 65.24 & 28.60 & 54.44 \\
StructRAG & 62.00 & 71.22 & 71.75 & 79.61 & 65.71 & 78.57 & 31.80 & 55.07 \\
PathRAG & 53.00 & 66.54 & 30.00 & 34.36 & 42.86 & 54.07 & 24.55 & 51.04 \\
MedRAG & 57.00 & 70.52 & 75.75 & 84.42 & 74.29 & \textbf{89.31} & 35.40 & 55.51 \\
MKRAG & 60.00 & 71.14 & 51.25 & 65.56 & 62.86 & 79.33 & 28.80 & 55.35 \\
\rowcolor{blue!5} CGR (ours) & \textbf{82.00} & \textbf{86.67} & \textbf{86.75} & \textbf{89.91} & \textbf{80.00} & 85.44 & \textbf{40.60} & \textbf{57.99} \\
\midrule
\multicolumn{9}{c}{\cellcolor{gray!10}\textit{Qwen2.5-14B-Instruct}} \\
\midrule
Zero-Shot & 34.00 & 39.04 & 43.20 & 51.14 & 34.29 & 43.09 & 15.80 & 33.44 \\
RAG & 44.00 & 52.01 & 45.50 & 55.54 & 42.86 & 50.19 & 31.80 & \textbf{47.81} \\
CoT & 49.00 & 53.50 & 42.50 & 52.44 & 37.14 & 46.24 & 20.80 & 34.88 \\
\rowcolor{blue!5} CGR (ours) & \textbf{55.00} & \textbf{64.47} & \textbf{61.30} & \textbf{67.40} & \textbf{51.43} & \textbf{54.11} & \textbf{34.60} & 44.08 \\
\midrule
\multicolumn{9}{c}{\cellcolor{gray!10}\textit{LLaMA-3.1-8B-Instruct}} \\
\midrule
Zero-Shot & 30.00 & 36.17 & 34.75 & 43.75 & 37.14 & 48.78 & 15.60 & 35.07 \\
RAG & 47.00 & 52.36 & 46.75 & 55.17 & 45.71 & 50.09 & 33.40 & \textbf{53.66} \\
CoT & 50.00 & \textbf{58.67} & 40.50 & 47.82 & 37.14 & 48.16 & 18.40 & 32.01 \\
\rowcolor{blue!5} CGR (ours) & \textbf{52.00} & 54.17 & \textbf{61.50} & \textbf{67.36} & \textbf{48.57} & \textbf{50.48} & \textbf{35.80} & 47.21 \\
\bottomrule
\end{tabular}
\end{table*}

\subsubsection{Baselines.}
We compare against three categories of methods:
\textbf{Non-retrieval:} Zero-shot prompting and chain-of-thought (CoT) reasoning \cite{wei2022chain};
\textbf{Standard RAG:} Dense retrieval with MedCPT \cite{jin2023medcpt}, followed by LLM generation;
\textbf{Graph-augmented RAG:} HippoRAG2 \cite{gutierrez2025rag}, StructRAG \cite{li2024structrag}, PathRAG \cite{chen2025pathrag}, HyperGraphRAG \cite{luo2025}, MedRAG \cite{medrag2024}, and MKRAG \cite{shi2025mkrag}.

\subsubsection{Evaluation Metrics.} We evaluate using Exact Match (EM) and token-level F1. EM measures whether the predicted answer exactly matches the ground truth after normalization. F1 computes the harmonic mean of precision and recall over normalized token overlap between prediction and ground truth, following prior work~\cite{lyu2024retrieve}.

\subsubsection{Implementation.}
All baseline methods use GPT-5.2 \cite{singh2025openai} for answer generation. For CGR, we use Qwen2.5-14B-Instruct-GPTQ-Int4 \cite{team2024qwen2} for tuple extraction (ablated in \S\ref{subsec:ablation}), deployed via vLLM \cite{kwon2025vllm} for optimized performance, and MedEmbed-large-v0.1 \cite{balachandran2024medembed} for path ranking. We adapt all methods to operate in a \textbf{post-retrieval setting}, where reasoning is performed exclusively over the provided gold documents. We leave integration with full scientific document retrieval \cite{yunyisemrank, kweon2025topic, kweon2026pairsem} as future work.

\subsection{Results}
\label{sec:results}

Table ~\ref{tab:main_results} presents our results across four biomedical QA benchmarks. CGR achieves state-of-the-art performance across all datasets and foundation models, with particularly strong gains on conditional questions.

\subsubsection{Main Findings.} With GPT-5.2, CGR achieves 82.00\% EM on CondMedQA, outperforming the strongest baseline by 20 points. On MedHopQA, CGR reaches 86.75\% EM compared to MedRAG's 75.75\%, an 11-point improvement. These gains are consistent across models: with Qwen2.5-14B, CGR improves over RAG by 11 points EM on CondMedQA and 15 points on MedHopQA.

\subsubsection{Analysis.} We highlight several key observations from our results below and analyze failure modes in Table~\ref{tab:error_analysis} of Appendix~\ref{app:error_analysis}.
\begin{enumerate}[label=(\arabic*), leftmargin=*, itemsep=2pt, topsep=2pt]
    \item \textit{Condition-dependent questions benefit most.} CGR shows its strongest and most consistent improvements on condition-dependent benchmarks (CondMedQA and MedHopQA-Cond), where constraints alter the correct answer and must be respected during reasoning.
    \item \textit{Strong performance on conditional and non-conditional queries.} CGR does not require questions to be conditional. When no conditions are present, edges remain ungated and traversal proceeds normally. This is reflected in strong performance on MedHopQA and BioASQ-B, which evaluate standard multi-hop questions.
    \item \textit{Consistent improvement across model scales.} CGR improves over baselines regardless of foundation model size, from GPT-5.2 to LLaMA-3.1-8B, showing that the structured reasoning framework provides value beyond what larger models alone can achieve.
    \item \textit{Graph-based methods not explicitly modeling conditionality underperform.} HippoRAG2, HyperGraphRAG, PathRAG, StructRAG, and our ablation study of CGR without gating (\S\ref{subsec:ablation}) all lag behind CGR, suggesting that naive graph construction without explicit condition modeling is insufficient for biomedical reasoning.
\end{enumerate}

\subsection{Ablation Studies}
\label{subsec:ablation}

We conduct ablation studies to validate the necessity of condition gating, extraction, and analyze sensitivity to key hyperparameters.

\begin{table}[h]
\centering
\caption{\normalfont Ablation study on the effect of condition gating.}
\label{tab:ab_gating}
\footnotesize
\setlength{\tabcolsep}{4pt}
\begin{tabular}{l|cc|cc|cc}
\toprule
\multirow{2}{*}{\textbf{Method}} 
& \multicolumn{2}{c|}{\textbf{CondMedQA}} 
& \multicolumn{2}{c|}{\textbf{MedHopQA}} 
& \multicolumn{2}{c}{\textbf{MedHop (C)}} \\
\cmidrule(lr){2-3} \cmidrule(lr){4-5} \cmidrule(lr){6-7}
 & EM & F1 & EM & F1 & EM & F1 \\
\midrule
CGR w/o gating & 57.0 & 60.2 & 72.5 & 76.8 & 62.9 & 73.5 \\
\rowcolor{blue!5} CGR & \textbf{82.0} & \textbf{86.7} & \textbf{86.8} & \textbf{89.9} & \textbf{80.0} & \textbf{85.4} \\
\bottomrule
\end{tabular}
\end{table}

\begin{figure*}[t]
\centering
\pgfplotsset{
    sensitivity/.style={
        width=0.32\textwidth,
        height=0.18\textwidth,
        xlabel style={font=\small},
        ylabel style={font=\small},
        tick label style={font=\footnotesize},
        title style={font=\small\bfseries},
        grid=major,
        grid style={gray!30, dashed, thin},
        mark size=2pt,
        every axis plot/.append style={line width=1pt},
        ymin=60, ymax=100,
    }
}
\begin{tikzpicture}
\begin{axis}[
    sensitivity,
    title={CondMedQA},
    xlabel={$k_{\text{paths}}$},
    ylabel={Score},
    xtick={1,3,5,10},
]
\addplot[color=orange, mark=square*] coordinates {(1,73) (3,82) (5,69) (10,70)};
\addplot[color=blue, mark=triangle*] coordinates {(1,76.61) (3,86.67) (5,71.78) (10,73.32)};
\end{axis}
\end{tikzpicture}%
\hfill
\begin{tikzpicture}
\begin{axis}[
    sensitivity,
    title={MedHopQA},
    xlabel={$k_{\text{paths}}$},
    xtick={1,3,5,10},
]
\addplot[color=orange, mark=square*] coordinates {(1,69.75) (3,86.75) (5,74.50) (10,78.75)};
\addplot[color=blue, mark=triangle*] coordinates {(1,74.11) (3,89.91) (5,79.29) (10,85.24)};
\end{axis}
\end{tikzpicture}%
\hfill
\begin{tikzpicture}
\begin{axis}[
    sensitivity,
    title={MedHopQA (Cond)},
    xlabel={$k_{\text{paths}}$},
    xtick={1,3,5,10},
]
\addplot[color=orange, mark=square*] coordinates {(1,71.42) (3,80.00) (5,85.71) (10,91.43)};
\addplot[color=blue, mark=triangle*] coordinates {(1,78.77) (3,85.44) (5,91.16) (10,96.87)};
\end{axis}
\end{tikzpicture}

\begin{tikzpicture}
\begin{axis}[
    sensitivity,
    xlabel={$k_{\text{nodes}}$},
    ylabel={Score},
    xtick={1,3,5,10},
]
\addplot[color=orange, mark=square*] coordinates {(1,67) (3,69) (5,82) (10,70)};
\addplot[color=blue, mark=triangle*] coordinates {(1,68.45) (3,72.14) (5,86.67) (10,72.45)};
\end{axis}
\end{tikzpicture}%
\hfill
\begin{tikzpicture}
\begin{axis}[
    sensitivity,
    xlabel={$k_{\text{nodes}}$},
    xtick={1,3,5,10},
]
\addplot[color=orange, mark=square*] coordinates {(1,72.25) (3,78.75) (5,86.75) (10,74.25)};
\addplot[color=blue, mark=triangle*] coordinates {(1,77.15) (3,82.38) (5,89.91) (10,79.55)};
\end{axis}
\end{tikzpicture}%
\hfill
\begin{tikzpicture}
\begin{axis}[
    sensitivity,
    xlabel={$k_{\text{nodes}}$},
    xtick={1,3,5,10},
]
\addplot[color=orange, mark=square*] coordinates {(1,74.28) (3,77.14) (5,80) (10,80)};
\addplot[color=blue, mark=triangle*] coordinates {(1,82.38) (3,82.32) (5,85.44) (10,85.44)};
\end{axis}
\end{tikzpicture}

\centerline{\small\textbf{Legend:}\quad \textcolor{orange}{\rule[0.3ex]{1.2em}{1.5pt}} EM \quad \textcolor{blue}{\rule[0.3ex]{1.2em}{1.5pt}} F1}
\caption{\normalfont Hyperparameter sensitivity analysis. Top row: varying $k_{\text{paths}} \in \{1, 3, 5, 10\}$. Bottom row: varying $k_{\text{nodes}} \in \{1, 3, 5, 10\}$. Based on these results, we set $k_{\text{paths}} = 3$ and $k_{\text{nodes}} = 5$ for all experiments.}
\label{fig:hp_sensitivity}
\end{figure*}
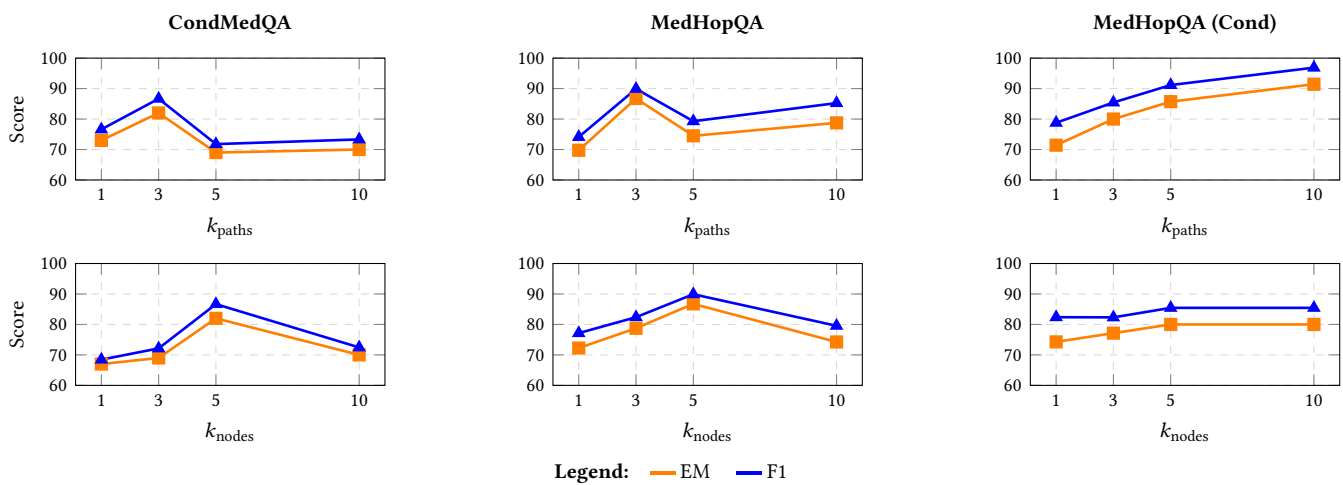

\subsubsection{Effect of Condition Gating.} To isolate the impact of our gating mechanism, we compare CGR to a variant that performs identical n-tuple extraction, graph construction, and traversal, but all edges are traversable regardless of context. As shown in Table~\ref{tab:ab_gating}, removing condition gating causes substantial performance degradation on CondMedQA and MedHopQA-Cond, while the drop on MedHopQA is more modest. This aligns with expectations, as CondMedQA and MedHopQA-Cond explicitly require conditional reasoning, making gating essential. MedHopQA tests multi-hop QA, so ungated traversal remains effective. These results confirm our hypothesis that condition gating is a critical reasoning component for context-aware QA.

\subsubsection{Effect of Extraction Model.} We ablate the LLM used for n-tuple graph extraction while keeping the rest of the pipeline fixed. We compare three extraction models: (1) Qwen2.5-14B-Instruct-GPTQ-Int4 \cite{team2024qwen2}, (2) Flan-T5-Large \cite{chung2024scaling}, and (3) LLaMA-3.2-3B-Instruct \cite{dubey2024llama}, all served locally via vLLM. For each configuration, we evaluate on a random 20-question subset (seed 42) from each benchmark; reasoning and answer generation use GPT-5.2. Flan-T5 uses shorter passage chunks (1500 chars) to respect its 512-token context limit. As shown in Table~\ref{tab:ab_extraction}, Qwen2.5-14B achieves the best F1 on MedHopQA (92.50) and ties for best on MedHopQA-Cond (88.33), while all models perform similarly on CondMedQA. LLaMA-3.2-3B matches Qwen2.5-14B on conditional benchmarks at comparable throughput (15.04 vs 15.08 s/doc). These results demonstrate an accuracy-efficiency tradeoff, with Qwen2.5-14B providing the optimal balance for our experiments.

\begin{table}[h]
\centering
\caption{\normalfont Ablation study on n-tuple extraction model, showing that higher quality n-tuple extraction leads to better CGR performance. Tok/s = tokens per second, s/doc = seconds per document (throughput from vLLM)}
\label{tab:ab_extraction}
\footnotesize
\setlength{\tabcolsep}{3pt}
\begin{tabular}{l|cc|cc|cc|c|c}
\toprule
\multirow{2}{*}{\textbf{Extractor}} 
& \multicolumn{2}{c|}{\textbf{CondMedQA}} 
& \multicolumn{2}{c|}{\textbf{MedHopQA}} 
& \multicolumn{2}{c|}{\textbf{MedHop (C)}} 
& \textbf{Tok/s} 
& \textbf{s/doc} \\
\cmidrule(lr){2-3} \cmidrule(lr){4-5} \cmidrule(lr){6-7}
 & EM & F1 & EM & F1 & EM & F1 &  &  \\
\midrule
Flan-T5-Large & 65.00 & 66.67 & 85.00 & 90.00 & 70.00 & 77.33 & 1053.95 & 21.69 \\
LLaMA-3.2-3B & 65.00 & 66.67 & 85.00 & 90.00 & 85.00 & 88.33 & 932.12 & \textbf{15.04} \\
\rowcolor{blue!5} Qwen2.5-14B & 65.00 & 66.67 & 85.00 & \textbf{92.50} & 85.00 & 88.33 & \textbf{1077.55} & 15.08 \\
\bottomrule
\end{tabular}
\end{table}

\subsubsection{Hyperparameter Sensitivity.} Figure~\ref{fig:hp_sensitivity} analyzes sensitivity across key hyperparameters. Top-k paths controls how many candidate reasoning paths are passed to the LLM for answer generation after ranking. Performance improves from $k{=}1$ to $k{=}3$ as additional paths provide supporting evidence, but degrades at $k{=}10$ due to noise from lower-ranked paths. Top-k nodes per keyword determines how many entry points are selected when initializing graph traversal; retrieving $k{=}5$ entry nodes provides sufficient coverage of relevant subgraphs without introducing irrelevant regions. CGR demonstrates stable performance across hyperparameter ranges, with consistent improvements over baselines in all settings.

\section{Interdisciplinary Contributions}

\noindent \textbf{Contributions to Biomedical Research.}
The CGR algorithm enables traceable conditional reasoning by constructing evidence trails grounded in published medical research.
Additionally, the CondMedQA benchmark addresses a critical gap by providing a dedicated evaluation dataset for conditional biomedical reasoning.

\smallskip

\noindent \textbf{Challenges Addressed by AI/ML.}
Current RAG systems produce answers without exposing the reasoning chain that connects evidence to conclusions.
CGR addresses this by structuring reasoning as explicit graph traversal grounded in the source text, rather than treating the model as a black box.

\smallskip

\noindent\textbf{Challenges of Using AI/ML.} Deploying AI/ML in clinical settings requires expert verification of outputs, adaptation to evolving guidelines, and human oversight for liability. \textsc{CGR's} interpretable reasoning paths address this gap, but AI recommendations are intended to assist clinical decision-making, not replace it.

\section{Limitations and Ethical Considerations}
\label{sec:limitations}
While CGR demonstrates strong performance on conditional biomedical reasoning, we acknowledge several limitations of our approach and important ethical considerations for deployment in clinical settings.

\subsection{Limitations}
\noindent\textbf{Benchmark Scope.} CondMedQA comprises 100 questions and serves as a diagnostic benchmark for conditional biomedical reasoning. Future work will expand coverage across condition types.

\smallskip
\noindent\textbf{Knowledge Source Dependence.} CGR constructs knowledge graphs from retrieved documents, inheriting any errors, omissions, or biases present in the source material.

\subsection{Ethical Considerations}

\noindent \textbf{Intended Use and Clinical Disclaimer.} CGR and CondMedQA are designed for research purposes only and are not intended for clinical decision-making. The system should not be used to provide medical advice, diagnosis, or treatment recommendations.

\smallskip

\noindent \textbf{Bias and Fairness.} Medical knowledge bases may reflect biases in clinical research, including underrepresentation of certain groups in clinical trials. CGR inherits these biases through its reliance on extracted medical knowledge. Auditing CondMedQA for demographic bias represents important future work.

\smallskip

\noindent \textbf{Data Privacy and Consent.} CondMedQA was constructed from publicly available medical literature and does not contain protected health information or data from human subjects. The LLM-assisted generation process used only synthetic clinical scenarios.

\section{Conclusion}
In this paper, we introduced CondMedQA, a benchmark for context-dependent biomedical question answering where patient-specific conditions change the correct answer.
We also propose \textsc{Condition-Gated Reasoning (CGR)}, a framework that explicitly models these conditional dependencies through structured knowledge graph traversal.
By extracting condition-aware n-tuples and gating graph edges based on context, CGR achieves state-of-the-art performance across multiple biomedical QA benchmarks, significantly outperforming existing approaches that treat all retrieved information uniformly. Our results demonstrate that explicit conditional modeling offers a promising paradigm for medical QA, addressing a fundamental limitation of current systems that often ignore contraindications, drug interactions, and other safety concerns.
We believe this approach can be extended to other high-stakes domains where contextual factors modify correct answers, offering a general framework for condition-aware reasoning.

\section*{Acknowledgements}
Research was supported in part by the AI Institute for Molecular Discovery, Synthetic Strategy, and Manufacturing: Molecule Maker Lab Institute (MMLI), funded by U.S. National Science Foundation under Award No.~2505932, NSF IIS 25-37827, and the Institute for Geospatial Understanding through an Integrative Discovery Environment (I-GUIDE) by NSF under Award No.~2118329. The research has used the Delta/DeltaAI advanced computing and data resource, supported in part by the University of Illinois Urbana-Champaign and through allocation \#250851 from the Advanced Cyberinfrastructure Coordination Ecosystem: Services \& Support (ACCESS) program, which is supported by National Science Foundation grants OAC 2320345, \#2138259, \#2138286, \#2138307, \#2137603, and \#2138296. Any opinions, findings, and conclusions or recommendations expressed herein are those of the authors and do not necessarily represent the views, either expressed or implied, of DARPA or the U.S. Government.

\bibliographystyle{ACM-Reference-Format}
\balance
\bibliography{reference}

\onecolumn 

\clearpage
\appendix
\raggedbottom

\section*{Appendix}
\bigskip
\section{Use of Large Language Models} 

In this work, large language models (LLMs) were used to assist with editing and refining of the manuscript. All research ideas, methodology, and experimental work were conducted by the authors.

\section{Extraction Examples}
\label{app:extraction_examples}

Table~\ref{tab:extraction_examples} presents how our extraction pipeline captures entities, relations, and patient-specific conditions that determine when relationships hold.

\begin{table*}[!h]
\centering
\caption{\normalfont Examples of condition-aware n-tuple extraction. \textcolor{answer}{Green text} indicates patient-specific conditions extracted from source text and captured in the n-tuple conditions field.}
\label{tab:extraction_examples}
\footnotesize
\renewcommand{\arraystretch}{1.15}
\begin{tabularx}{\textwidth}{X}
\toprule
\textbf{Document 1: HIV Treatment} \\
\midrule
\textit{``A large study in Africa and India found that a PI-based regimen was superior to an NNRTI-based regimen \textcolor{answer}{in children less than 3 years} who had \textcolor{answer}{never been exposed to NNRTIs} in the past. Thus the WHO recommends PI-based regimens \textcolor{answer}{for children less than 3}. \textcolor{answer}{In pregnant women}, viral load is proportional to transmission risk; ART reduces transmission to mothers and infants \textcolor{answer}{before, during, and after delivery}.''} \\[4pt]
\textbf{Extracted N-Tuples:} \\
$\langle$PI-based regimen, \textsc{superior\_to}, NNRTI-based regimen, [\textcolor{answer}{children \textless 3 years}, \textcolor{answer}{not exposed to NNRTI}]$\rangle$ \\
$\langle$WHO, \textsc{recommends}, PI-based regimen, [\textcolor{answer}{children \textless 3 years}]$\rangle$ \\
$\langle$viral load, \textsc{proportional\_to}, transmission risk, [\textcolor{answer}{pregnant women}]$\rangle$ \\
$\langle$ART, \textsc{reduces}, transmission risk, [\textcolor{answer}{pregnant women}, \textcolor{answer}{before/during/after delivery}]$\rangle$ \\
\midrule
\textbf{Document 2: Misoprostol} \\
\midrule
\textit{``Misoprostol should not be taken by \textcolor{answer}{pregnant women} with wanted pregnancies to reduce the risk of NSAID-induced gastric ulcers because it increases uterine tone and contractions \textcolor{answer}{in pregnancy}, which may cause partial or complete abortions, and because its use \textcolor{answer}{in pregnancy} has been associated with birth defects.''} \\[4pt]
\textbf{Extracted N-Tuples:} \\
$\langle$misoprostol, \textsc{contraindicated\_in}, wanted pregnancies, [\textcolor{answer}{pregnant women}]$\rangle$ \\
$\langle$misoprostol, \textsc{increases}, uterine tone and contractions, [\textcolor{answer}{pregnancy}]$\rangle$ \\
$\langle$misoprostol, \textsc{causes}, partial or complete abortion, [\textcolor{answer}{pregnancy}]$\rangle$ \\
$\langle$misoprostol, \textsc{associated\_with}, birth defects, [\textcolor{answer}{pregnancy}]$\rangle$ \\
\midrule
\textbf{Document 3: Urinary Tract Infection} \\
\midrule
\textit{``Urinary tract symptoms are frequently lacking \textcolor{answer}{in the elderly}. The presentations may be vague and include incontinence, a change in mental status, or fatigue as the only symptoms. \textcolor{answer}{In young healthy women}, diagnosis can be made from symptoms alone. \textcolor{answer}{Postmenopausal women} may be treated with vaginal estrogen replacement. Risk factors include sexual activity \textcolor{answer}{in young women}, chronic prostatitis \textcolor{answer}{in males}, and vesico-ureteral reflux \textcolor{answer}{in children}.''} \\[4pt]
\textbf{Extracted N-Tuples:} \\
$\langle$UTI, \textsc{diagnosed\_by}, symptoms alone, [\textcolor{answer}{young healthy women}]$\rangle$ \\
$\langle$UTI, \textsc{treated\_with}, vaginal estrogen, [\textcolor{answer}{postmenopausal women}]$\rangle$ \\
$\langle$UTI, \textsc{presents\_with}, incontinence/mental status change/fatigue, [\textcolor{answer}{elderly}]$\rangle$ \\
$\langle$UTI, \textsc{risk\_factor}, sexual activity, [\textcolor{answer}{young women}]$\rangle$ \\
$\langle$UTI, \textsc{risk\_factor}, chronic prostatitis, [\textcolor{answer}{males}]$\rangle$ \\
$\langle$UTI, \textsc{risk\_factor}, vesico-ureteral reflux, [\textcolor{answer}{children}]$\rangle$ \\
\bottomrule
\end{tabularx}
\end{table*}

\pagebreak 

\section{Query Parsing Examples}
\label{app:query_parsing}

Table~\ref{tab:query_parsing_examples} demonstrates condition gating and entity negation.
Query parsing transforms natural-language questions into structured representations for graph matching and condition gating. We define three components:
\begin{itemize}[leftmargin=2em, itemsep=2pt, topsep=2pt]
    \item \textbf{K} (Keywords): Entities and concepts the answer must match; drives node retrieval in the graph
    \item \textbf{C} (Conditions): Required context that must hold (e.g., ``in children'') and excluded context that must not hold (e.g., ``not in adults''); used for edge gating
    \item \textbf{N} (Negated Entities): Entities explicitly excluded as answers (e.g., ``distinct from X'', ``other than Y'')
\end{itemize}

\begin{table*}[!htbp]
\centering
\caption{\normalfont Query parsing examples showing decomposition into keywords (\textbf{K}), conditions (\textbf{C}), and negated entities (\textbf{N}).}
\label{tab:query_parsing_examples}
\small
\renewcommand{\arraystretch}{1.2}
\begin{tabularx}{\textwidth}{lX}
\toprule
\multicolumn{2}{l}{\textbf{Example 1:} \textit{``Which gene causes cardiomyopathy in pediatric patients but not in adults?''}} \\
\midrule
\textbf{K} (keywords) & \texttt{[gene, cardiomyopathy, causes]} \\
\textbf{C} (required) & \texttt{[pediatric, in children]} \\
\textbf{C} (excluded) & \texttt{[in adults, adult patients]} \\
\textbf{N} (negated) & \texttt{[]} \\
\midrule
\multicolumn{2}{p{\textwidth}}{\textit{Interpretation:} \textbf{K} matches nodes related to genes and cardiomyopathy. \textbf{C} gates traversal: only edges with pediatric/child conditions are followed; edges conditioned on adults are blocked.} \\
\midrule
\midrule
\multicolumn{2}{l}{\textbf{Example 2:} \textit{``What drug treats hypertension in pregnant women, excluding ACE inhibitors?''}} \\
\midrule
\textbf{K} (keywords) & \texttt{[drug, hypertension, treats]} \\
\textbf{C} (required) & \texttt{[in pregnancy, pregnant women]} \\
\textbf{C} (excluded) & \texttt{[]} \\
\textbf{N} (negated) & \texttt{[ACE inhibitors]} \\
\midrule
\multicolumn{2}{p{\textwidth}}{\textit{Interpretation:} \textbf{K} retrieves drug and hypertension nodes. \textbf{C} restricts traversal to pregnancy-safe edges. \textbf{N} filters the final answer set: even if ACE inhibitors appear in valid paths, they are excluded from the response.} \\
\bottomrule
\end{tabularx}
\end{table*}

\section{Condition Evaluation Examples}
\label{app:condition_eval}

Table~\ref{tab:condition_eval_examples} presents the examples of the condition evaluation.
The condition evaluation step uses a single LLM call to populate the lookup table $\mathcal{L}$ for edge gating, enabling multi-hop reasoning with each edge subject to condition gating.
Given a query and a set of conditions extracted from graph edges, the LLM evaluates whether each condition is satisfied (\texttt{true}), violated (\texttt{false}), or unknown (\texttt{null}) in the query context.

\begin{table*}[!htbp]
\centering
\caption{\normalfont Condition evaluation examples showing how a single LLM call populates the lookup table $\mathcal{L}$ for edge gating.}
\label{tab:condition_eval_examples}
\small
\renewcommand{\arraystretch}{1.2}
\begin{tabularx}{\textwidth}{lX}
\toprule
\multicolumn{2}{l}{\textbf{Example 1:} \textit{``Which gene causes cardiomyopathy in pediatric patients but not in adults?''}} \\
\midrule
\textbf{Conditions} & \texttt{[pediatric, in children, in adults, adult patients]} \\
\midrule
\textbf{Output} & \texttt{\{``pediatric'': true, ``in children'': true, ``in adults'': false, ``adult patients'': false\}} \\
\midrule
\multicolumn{2}{p{\textwidth}}{\textit{Gating result:} Edges conditioned on ``pediatric'' or ``in children'' are traversable ($\mathcal{G}=1$); edges conditioned on ``in adults'' or ``adult patients'' are blocked ($\mathcal{G}=0$).} \\
\midrule
\midrule
\multicolumn{2}{l}{\textbf{Example 2:} \textit{``What antibiotic is safe for a 5-year-old boy with pneumonia and no known allergies?''}} \\
\midrule
\textbf{Conditions} & \texttt{[in children, in adults, penicillin allergy, renal impairment, pregnancy]} \\
\midrule
\textbf{Output} & \texttt{\{``in children'': true, ``in adults'': false, ``penicillin allergy'': false, ``renal impairment'': null, ``pregnancy'': false\}} \\
\midrule
\multicolumn{2}{p{\textwidth}}{\textit{Gating result:} Edges requiring ``in children'' are traversable. Edges conditioned on ``penicillin allergy'' remain open (patient has no allergy, so condition is not violated). ``renal impairment'' evaluates to \texttt{null} (unknown), so those edges remain traversable. Edges requiring ``pregnancy'' or ``in adults'' are blocked.} \\
\bottomrule
\end{tabularx}
\end{table*}



\section{Error Analysis}
\label{app:error_analysis}

We manually analyze the 18 incorrect CGR predictions on CondMedQA and categorize them into three error types: retrieval (11), extraction/normalization (5), and reasoning (2). Table~\ref{tab:error_analysis} presents two representative examples per category; we discuss each below.

\begin{table*}[ht!]
\small
\caption{\normalfont Error analysis of 18 incorrect CGR predictions on CondMedQA.
Patient-specific conditions are highlighted in \textcolor{answer}{color}.
Two representative examples are shown per error type.}
\setlength{\tabcolsep}{4pt}
\renewcommand{\arraystretch}{1.05}
\begin{tabularx}{\textwidth}{
    p{1.8cm}
    p{0.6cm}
    X
    p{1.4cm}
    p{1.6cm}
    X
}
\toprule
\textbf{Error Type} & \textbf{QID} & \textbf{Question} &
\textbf{Gold} & \textbf{Predicted} & \textbf{Diagnosis} \\
\midrule
\textbf{Retrieval} \newline (11/18)
& 23
& What antibiotic is recommended for treating scrub typhus
\textcolor{answer}{during pregnancy}?
& Azithromycin
& Doxycycline
& Gold tuple with \textcolor{answer}{pregnancy} condition exists in KG but not in top-ranked paths; typhus--doxycycline edge dominates ranking. \\
\cmidrule{2-6}
& 74
& Which multikinase inhibitor is most effective for hepatocellular carcinoma
\textcolor{answer}{in advanced (BCLC-C) unresectable disease}?
& Sorafenib
& Insufficient evidence
& Gold tuple with \textcolor{answer}{BCLC-C} condition in KG; top paths traverse unrelated edges. Gold never reaches evidence assembly. \\
\midrule
\textbf{Extraction} \newline (5/18)
& 65
& Which csDMARD that has the side effect of liver cancer is the most effective for ankylosing spondylitis
\textcolor{answer}{in peripheral arthritis}?
& Methotrexate
& Insufficient evidence
& Methotrexate not extracted from source documents; KG lacks the gold entity entirely. \\
\cmidrule{2-6}
& 96
& Which anxiolytic is preferred for generalized anxiety disorder
\textcolor{answer}{in a patient who has myasthenia gravis}?
& Buspirone
& Sertraline
& Buspirone absent from KG. Model correctly avoids benzodiazepines and selects SSRI; normalization maps SSRI to an incorrect canonical form. \\
\midrule
\textbf{Reasoning} \newline (2/18)
& 28
& What imaging test is preferred for suspected pulmonary embolism
\textcolor{answer}{in a pregnant patient} due to lower radiation exposure?
& V/Q scan
& MRI w/o contrast
& V/Q is the guideline alternative to CTPA in pregnancy; model over-optimizes for radiation avoidance and selects MRI. \\
\cmidrule{2-6}
& 99
& Which antimycobacterial drug should be replaced in the RIPE regimen for
\textcolor{answer}{a patient who is HIV-positive and receiving protease inhibitors}?
& Rifabutin
& Rifampin
& Gold present in retrieved paths. Model answers the drug \emph{to be} replaced (rifampin) rather than the replacement (rifabutin). \\
\bottomrule
\end{tabularx}
\label{tab:error_analysis}
\end{table*}

\paragraph{Retrieval errors.} The gold answer entity is present in the constructed knowledge graph, but path ranking does not surface it in the top-$k$ paths passed to the answer generation model. In Q23, the query asks for an antibiotic for scrub typhus \emph{during pregnancy}. The knowledge graph contains the conditional tuple $\langle$scrub typhus, treated\_with, azithromycin, [pregnancy]$\rangle$, correctly encoding that azithromycin is the pregnancy-safe alternative. However, the top-ranked paths instead traverse the more heavily connected typhus--doxycycline edge, which dominates due to higher semantic similarity with the query entity ``scrub typhus.'' The gating mechanism correctly annotates doxycycline edges with pregnancy contraindications, but because azithromycin paths rank below the top-$k$ cutoff, the model never sees the correct alternative. In Q74, a similar pattern emerges: the tuple $\langle$liver carcinoma, treated\_by, sorafenib, [BCLC stage C]$\rangle$ exists with the exact condition matching the query, yet the top paths traverse tangentially related edges (e.g., liver carcinoma $\rightarrow$ ethanol, liver carcinoma $\rightarrow$ hepatitis B), and the model concludes ``insufficient evidence'' because sorafenib never appears in the assembled evidence. These cases reveal that while CGR's edge gating effectively blocks contraindicated paths, incorporating condition-match signals into the ranking score (Eq.~\ref{eq:path-ranking}) is a promising direction for reducing retrieval errors.

\paragraph{Extraction and normalization errors.} In these cases, the gold entity is entirely absent from the knowledge graph. In Q65, the query asks for a csDMARD for ankylosing spondylitis in peripheral arthritis, with the gold answer being methotrexate. However, the extraction model produces tuples mentioning only glucocorticoids and TNF inhibitors from the source documents. The model reasonably responds ``insufficient evidence,'' which is correct given its available knowledge but wrong against the benchmark. In Q96, the query asks for an anxiolytic for generalized anxiety disorder in a patient with myasthenia gravis (gold: buspirone). Buspirone does not appear in the extracted tuples. The model demonstrates sound conditional reasoning by correctly avoiding benzodiazepines (contraindicated in myasthenia gravis) and selecting an SSRI instead, but chooses sertraline rather than the gold answer. Additionally, the UMLS-based entity normalization maps the predicted SSRI to ``SsrI endonuclease,'' a restriction enzyme, illustrating how resolution errors can compound extraction gaps. These cases suggest that improving extraction recall and auditing the UMLS normalization pipeline are important future directions.

\paragraph{Reasoning errors.} These occur when the gold entity is present in the retrieved evidence but the answer generation model selects an incorrect response. In Q28, the query asks for the preferred imaging test for suspected pulmonary embolism in a pregnant patient with lower radiation exposure. The benchmark answer is a V/Q scan, which delivers less radiation than CT pulmonary angiography. The model instead selects MRI without contrast, which avoids ionizing radiation entirely but is not the standard recommendation due to limited availability and lower sensitivity. This suggests the model over-optimized for the ``lower radiation'' constraint without grounding its choice in clinical guideline knowledge. In Q99, the query asks which antimycobacterial drug should be replaced in the RIPE regimen for an HIV-positive patient on protease inhibitors. The gold answer is rifabutin (the replacement drug), but the model answers rifampin (the drug to be replaced). Notably, rifabutin appears in the retrieved reasoning paths with the correct HIV/antiretroviral conditions, so the model had access to the necessary information and just failed to synthesize the reasoning traces.

\section{Prompts}
\label{app:prompts}

This appendix provides the full prompts used in the CGR pipeline: query parsing (\S\ref{app:query_prompt}), knowledge extraction (\S\ref{app:extraction_prompt}), condition evaluation (\S\ref{app:condition_prompt}), and answer generation (\S\ref{app:answer_prompt}).

\subsection{Query Parsing Prompt}
\label{app:query_prompt}

This prompt parses natural language questions into structured intent representations, separating target entities, required attributes, negations, and patient-specific conditions to enable condition-aware graph traversal.

\begin{tcolorbox}[title=Query Parsing Prompt, colback=blue!5, colframe=blue!50!black, fonttitle=\bfseries]
\small
\textbf{System:} You are a query parser for biomedical knowledge graphs. Parse the question into a structured intent-aware format.

\textbf{Schema:}
\begin{itemize}[leftmargin=*, itemsep=0pt, topsep=2pt]
    \item \texttt{target\_entity}: The main concept being asked about
    \item \texttt{positive\_attributes}: Properties the answer MUST have (2--5 items)
    \item \texttt{negated\_entities}: Entities explicitly mentioned as NOT the answer (e.g., ``distinct from'', ``other than'')
    \item \texttt{required\_conditions}: Conditions that MUST hold (e.g., ``in males'', ``during pregnancy'')
    \item \texttt{excluded\_conditions}: Conditions that must NOT be present
\end{itemize}

\textbf{Output:} JSON object with fields: \texttt{target\_type}, \texttt{target\_entity}, \texttt{positive\_attributes}, \texttt{negated\_entities}, \texttt{required\_conditions}, \texttt{excluded\_conditions}

\tcblower
\textbf{Examples:}
\begin{itemize}[leftmargin=*, itemsep=1pt, topsep=2pt]
    \item \textit{``Which antibody treats metastatic colorectal cancer?''} \\
    $\rightarrow$ \texttt{target\_type}: antibody, \texttt{positive\_attributes}: [antibody, treats cancer, metastatic colorectal cancer]
    \item \textit{``Which gene causes cardiomyopathy in pediatric patients but not in adults?''} \\
    $\rightarrow$ \texttt{required\_conditions}: [pediatric], \texttt{excluded\_conditions}: [in adults]
    \item \textit{``What drug treats hypertension in pregnant women, excluding ACE inhibitors?''} \\
    $\rightarrow$ \texttt{negated\_entities}: [ACE inhibitors], \texttt{required\_conditions}: [in pregnancy]
\end{itemize}
\end{tcolorbox}

\subsection{Knowledge Extraction Prompt}
\label{app:extraction_prompt}

This prompt extracts condition-aware knowledge graph n-tuples from biomedical text, capturing entities, relations, and contextual qualifiers (e.g., ``in the liver'', ``during pregnancy'') that determine when relationships hold.

\begin{tcolorbox}[title=Knowledge Extraction Prompt, colback=green!5, colframe=green!50!black, fonttitle=\bfseries]
\small
\textbf{System:} You are a high-precision knowledge extraction engine for biomedical literature. Extract a dense, exhaustive knowledge graph from the passage as a JSON array of n-tuples with contextual conditions.

\textbf{Core Principles:}
\begin{itemize}[leftmargin=*, itemsep=0pt, topsep=2pt]
    \item \textit{Exhaustive Recall}: Capture every factual claim, no matter how small
    \item \textit{Atomicity}: Each n-tuple has exactly one subject and one object; split conjunctions
    \item \textit{Biomedical Anchoring}: Extract canonical entities; move descriptors (``elevated'', ``serum'') to conditions
    \item \textit{Context-Independent}: Extract from exclusions, differentials, and comparisons
\end{itemize}

\textbf{JSON Schema:} \texttt{entity1} (subject), \texttt{relation}, \texttt{inverse\_relation}, \texttt{entity2} (object), \texttt{conditions} (list of qualifiers)

\tcblower
\textbf{Example:}

\textit{Input:} ``L-type and T-type calcium channels are both blocked by Compound 99 in cardiomyocytes.''

\textit{Output:}
\begin{verbatim}
[{``entity1'': ``L-type calcium channels'', 
  ``relation'': ``blocked_by'', ``entity2'': ``Compound 99'',
  ``conditions'': [``in cardiomyocytes'']},
 {``entity1'': ``T-type calcium channels'', 
  ``relation'': ``blocked_by'', ``entity2'': ``Compound 99'',
  ``conditions'': [``in cardiomyocytes'']}]
\end{verbatim}
\end{tcolorbox}

\subsection{Condition Evaluation Prompt}
\label{app:condition_prompt}

This prompt evaluates whether patient-specific conditions mentioned in graph edges are satisfied, violated, or unknown given the query context, enabling the gating mechanism to block or permit edge traversal.

\begin{tcolorbox}[title=Condition Evaluation Prompt, colback=orange!5, colframe=orange!50!black, fonttitle=\bfseries]
\small
\textbf{System:} You are evaluating medical/clinical conditions against a query.

\textbf{Task:} For each condition extracted from the knowledge graph, determine if the patient/context in the query SATISFIES that condition.

\textbf{Return Values:}
\begin{itemize}[leftmargin=*, itemsep=0pt, topsep=2pt]
    \item \texttt{true}: Query explicitly or implicitly indicates condition IS satisfied
    \item \texttt{false}: Query explicitly or implicitly indicates condition is NOT satisfied
    \item \texttt{null}: Query does not mention anything relevant (unknown)
\end{itemize}

\textbf{Guidelines:}
\begin{itemize}[leftmargin=*, itemsep=0pt, topsep=2pt]
    \item Handle synonyms: ``in boys'' $\equiv$ ``male children'' $\equiv$ ``pediatric males''
    \item Handle numeric reasoning: ``5-year-old'' $\Rightarrow$ ``in adults'' is \texttt{false}, ``in children'' is \texttt{true}
    \item Handle implications: ``pregnant woman'' $\Rightarrow$ ``during pregnancy'' is \texttt{true}
    \item Handle negations: ``no kidney disease'' $\Rightarrow$ ``renal impairment'' is \texttt{false}
    \item When truly unknown, return \texttt{null}---do not guess
\end{itemize}

\tcblower
\textbf{Example:}

\textit{Query:} ``What antibiotic for a 5-year-old boy with pneumonia?''

\textit{Conditions:} [``in boys'', ``in adults'', ``eGFR < 30'']

\textit{Output:} \texttt{\{``in boys'': true, ``in adults'': false, ``eGFR < 30'': null\}}
\end{tcolorbox}

\subsection{Answer Generation Prompt}
\label{app:answer_prompt}

This prompt generates the final answer from retrieved evidence paths, instructing the model to select the best available option based on the gated graph traversal results rather than demanding perfect proof.

\begin{tcolorbox}[title=Answer Generation Prompt, colback=purple!5, colframe=purple!50!black, fonttitle=\bfseries]
\small
\textbf{System:} You are a biomedical question-answering assistant. Provide the BEST AVAILABLE ANSWER based on the evidence.

\textbf{Core Principle---Best Available Choice:}
\begin{itemize}[leftmargin=*, itemsep=0pt, topsep=2pt]
    \item If ONE option has relevant evidence and others don't $\rightarrow$ choose that option
    \item Warnings/cautions about a treatment do NOT disqualify it---they show it IS used
    \item Contraindications for OTHER options strengthen the case for the remaining option
    \item ``Insufficient evidence'' is a LAST RESORT---only when NO option has ANY relevant connection
\end{itemize}

\textbf{Response Format:}
\begin{itemize}[leftmargin=*, itemsep=0pt, topsep=2pt]
    \item \texttt{REASONING}: 2--4 sentences analyzing evidence with citations (e.g., [doc1])
    \item \texttt{ANSWER}: Single entity name, yes/no, or short phrase---NO full sentences
\end{itemize}

\textbf{Answer Formatting:}
\begin{itemize}[leftmargin=*, itemsep=0pt, topsep=2pt]
    \item Use what a clinician would say on rounds (e.g., ``MRI'' not ``Magnetic Resonance Imaging'')
    \item Be maximally concise: single word or shortest standard term
    \item Prefer clinical acronyms over spelled-out formal names
\end{itemize}
\end{tcolorbox}

\end{document}